\documentclass[runningheads]{llncs}
\usepackage{graphicx}
\usepackage{comment}
\usepackage{appendix}
\usepackage[nolist]{acronym}
\usepackage{multirow}
\usepackage{color}

\begin{document}
\title{Enriching the Machine Learning Workloads in BigBench}

\subtitle{Technical Report, May 2024}
%
%
\author{Matthias Polag, 
Todor Ivanov, and
Timo Eichhorn 
}
\authorrunning{M. Polag et al.}
%
\institute{Frankfurt Big Data Lab, Goethe University Frankfurt, Frankfurt, Hessen, Germany\\
\email{{matthias,todor,timo}@dbis.cs.uni-frankfurt.de}
}
\maketitle              
\begin{abstract}

In the era of Big Data and the growing support for Machine Learning, Deep Learning and Artificial Intelligence algorithms in the current software systems, there is an urgent need of standardized application benchmarks that stress test and evaluate these new technologies. Relying on the standardized BigBench (TPCx-BB) benchmark, this work enriches the improved BigBench V2 with three new workloads and expands the coverage of machine learning algorithms. Our workloads utilize multiple algorithms and compare different implementations for the same algorithm across several popular libraries like MLlib, SystemML, Scikit-learn and Pandas, demonstrating the relevance and usability of our benchmark extension.

\end{abstract}
\section{Introduction} \label{intro}
In current times \textit{Big Data} is often seen as the \textit{next big thing in innovation} \cite{gobble2013big} and has the potential to generate significant growth for the economy as a whole \cite{manyika2011big}. For example, the article \textit{Smart analytics: How marketing drives short-term and long-term growth} \cite{perrey2013smart} claims that big data analysis can increase the return of investment in retailer marketing by 15 to 20 percent.

Today there exist several frameworks designed to process big data like Hadoop \cite{white2012hadoop} and Spark \cite{zaharia2016apache}, yet it can be challenging to decide which system best meets their needs. To help with that decision end-to-end application benchmarks like BigBench (TPCx-BB) \cite{ghazal2013bigbench,TPCx-BB} were created. They evaluate different Big Data systems on a set of tasks and measure their capabilities. While BigBench and BigBench V2 \cite{ghazal2017bigbench} currently cover many common business tasks, they have only few examples of machine learning tasks as part of their workload and solely rely on the Mahout and partially MLlib libraries to implement the necessary algorithms. As suggested by Singh \cite{singh2016benchmarking} and briefly described in the vision of ABench \cite{IvanovS18}, we enrich BigBench with new machine learning algorithms and workloads, which then help us to evaluate and compare various popular machine learning libraries.

In this paper we provide an extension of BigBench V2 by adding additional machine learning workloads and algorithms using different libraries. 
The main contributions of this work are:
\begin{itemize}
    \item Enriching BigBench V2 with more machine learning algorithms.
    \item Implementing the workloads in several different libraries (MLlib, SystemML, Scikit-learn and Pandas).
    \item Comparison and evaluation of the libraries.
\end{itemize}
The paper is structured as follows: Section \ref{relatedwork} gives an overview of related work. Section \ref{machinelearning} starts with a brief overview of existing Machine Learning algorithms and libraries and introduces the BigBench Machine Learning workloads and their implementations. Section \ref{results} describes the experimental setup, methodology and execution, and presents the results of the experimental evaluation. Finally, Section \ref{confw} summarizes the lessons learned.

\section{Related Work} \label{relatedwork}

\subsubsection{Machine Learning (ML) benchmarks} are mainly realized as micro benchmarks. They focus on measuring the performance of specific components or tasks.

DeepBench \cite{DeepBench} measures the performance of running basic neural network operations regarding the used hardware and covers essential algorithms like \ac{GEMM}, Convolution, and \ac{RNN}. The results are adequate indicators of the execution time necessary to train an entire model.

DawnBench \cite{coleman2017dawnbench} uses image classification and question answering workloads to evaluate deep learning systems. It proposes a new metric, using time per accuracy as an indicator. The metric reports both the end-to-end training time to achieve a state-of-the-art accuracy together with the inference of that accuracy.

MLPerf \cite{MLperf} provides a benchmark suite to evaluate ML software frameworks, ML hardware accelerators, and ML cloud platforms. It measures the performance for training and inference by a mixture of image classification, object detection, translation, recommendation, and reinforcement learning workloads.

SparkBench \cite{Li:2015:SCB:2742854.2747283} is a benchmark suite consisting of different types of Spark workloads. The workloads can be categorized into Graph Computation, Machine Learning, SQL Query, and Streaming Application. The ML workloads cover Logistic Regression, \ac{SVM} and matrix factorization, whereas the SQL queries focus on SELECT, AGGREGATE and JOIN. 

BigDataBench \cite{wang2014bigdatabench} provides an extensive benchmark suite, covering structured, semi-structured and unstructured data sets. Various micro-benchmarks are implemented for the application domains. The ML performance measurements use Cifar and ImageNet data to execute the different Machine Learning workloads.

However, all of the above are micro-benchmarks or suites of micro-benchmarks targeting to measure particular functionality and capability of libraries. They cover many different ML algorithms, measure important ML metrics like accuracy, but do not provide a realistic use case where multiple libraries interface together in an application workflow (pipeline). This problem is addressed in the end-to-end application benchmarks simulating real industry scenarios.

\subsubsection{Big Data application-level benchmarks} concentrate on the entire system by performing common application scenarios \cite{han2018benchmarking}.


BigBench (TPCx-BB) \cite{TPCx-BB,ghazal2013bigbench} is a technology agnostic business-oriented end-to-end big data benchmark. It is modeled as a product retailer, containing information about selling different products to customers in physical as well as online stores. Based on its model it incorporates three different types of data sets - unstructured, semi-structured and structured data to cover the big data variety characteristics completely. The unstructured data consists of product reviews, the semi-structured is represented by weblogs, and the structured data is adapted from the TPC-DS \cite{TPC-DS} benchmark. All the data is synthetic and produced by a data generator, that is able to create data sets of different sizes defined by a \ac{SF}. The benchmark workloads consist of 30 different queries chosen to cover all aspects of big data analysis described in the McKinsey’s report \cite{manyika2011big}, with 10 of the queries taken from TPC-DS. 9 of these 30 queries focus on natural language processing or machine learning problems \cite{ghazal2013bigbench}.

In BigBench V2 \cite{ghazal2017bigbench} improvements are released to overcome the limitations of the initial BigBench benchmark. The schema and the resulting queries are adapted to more real-life situations using a star-schema, and also late binding is applied by using the key-value weblogs of the semi-structured data. Apart from that, it retains the same metrics and computations as BigBench. BigBench V2's workload consists of 30 queries, 17 of them were directly adopted from the initial BigBench while 13 new queries focusing on the semi-structured data were introduced.

A recent SPEC Big Data Research Group survey \cite{IvanovRPQPPB15} provides an extensive summary of the existing and currently under development Big Data benchmarks to help both researchers and practitioners choose the appropriate benchmark for their needs.

\section{BigBench Machine Learning Extension} \label{machinelearning}

\subsection{Machine Learning Background}
Machine learning algorithms can learn the underlying patterns in a data set without the creation of a specific algorithm. Based on the training data set, which needs to be provided for the algorithm, it can create a model containing the structural information of the data and use it to make predictions for unobserved data.There are several types of machine learning algorithms based on the tasks they are used for. The algorithms on which we focus in this paper can be classified into the following categories:
\paragraph{Classification:} This type of algorithm tries to separate data into several predefined classes based on their features. The goal of this algorithm is to predict to which class a new data point will belong. As an example classification algorithms are typically used for image recognition tasks.
\paragraph{Clustering:} Similar to classification, clustering also tries to group data into several classes. In distinction to classification for clustering the number of groups are undefined for the algorithm. Dividing customers into different groups to target them more precisely is a common clustering job.
\paragraph{Frequent pattern mining:} Is the discovery of item sets which frequently occur together in a list of transactions and further the process of generating association rules based on the found patterns. Frequent pattern mining is often used in market basket analysis to give insight into the behavior of shoppers. An example of an association rule in this context would be shoppers who bought toothbrushes will also purchase toothpaste.
\paragraph{Regression:} The relationship between several independent variables on a dependent variable is modeled, to predict the general behavior of the dependent variable. A basic example would be to predict the number of ice cream sales on any day based on the temperature and price of the ice cream.
\paragraph{Topic modeling:} Is often used for text-mining purposes. This class of machine learning algorithms identifies topics in a set of documents. Each topic is defined as a collection of words and each document is a mixture of topics.

For the sake of simplicity in this paper we focus on the following categories and algorithms:
\begin{table}
    \centering
    \caption{Overview of used Machine Learning classes and algorithms}
    \begin{tabular}{|c|p{29em}|}
        \hline
        Class & Algorithms \\
        \hline
        Classification & Naive Bayes, Logistic Regression, \ac{SVM}, \ac{MLP}, Decision Tree \\
        \hline
        Clustering & K-Mean, \ac{GMM} \\
        \hline
        Frequent Pattern & Frequent Pattern-Growth, Eclat \\
        \hline
        Topic Modeling & Latent Dirichlet Allocation (LDA) \\
        \hline
    \end{tabular}
    \label{MLclassAlgo}
\end{table}

\subsection{Technology Libraries}
To implement the new workloads, we use several machine learning libraries. For a better understanding, we provide some information on each of them.

\subsubsection{MLlib} \cite{SparkMLlib} is a machine learning library for Apache Spark, providing implementations for many state-of-the-art algorithms. MLlib supports Spark API and operates with Python and R.

\subsubsection{Scikit-learn and Pandas}
Pandas \cite{mckinney2011pandas} is a Python library for data analysis able to perform machine learning with the additional implementation of Scikit-learn \cite{pedregosa2011scikit}. Scikit-learn implements a vast variety of different machine learning algorithms, covering most algorithms in MLlib. We use Pandas and Scikit-learn to determine which amount of data is needed for parallel algorithms running on Spark to outperform Scikit-learn on a single machine. 

\subsubsection{Spark-fim} \cite{Sparkfim} offers the support to execute a distributed implementation of Frequent Itemsets Mining algorithms. It provides Eclat and BigFIM to run on top of Apache Spark \cite{Sparkfim}.

\subsubsection{SystemML} \cite{boehm2016systemml} provides many machine learning algorithms. It can be run as a standalone or deployed on Hadoop or Spark \cite{ghoting2011systemml,boehm2016systemml}. Unlike Spark or Python libraries like Pandas and Scikit-learn, it uses the matrix market format as input and not a dataframe object. Since there are no native possibilities to directly convert data stored in Hive into the desired input format, extensive formatting is required.

\subsection{Workload Description}\label{existingWL}
Five BigBench/BigBench V2 queries (Q05, Q20, Q25, Q26 and Q28) cover three Machine Learning algorithms (ML) in Clustering (K-Means) and Classification (Logistic Regression and Naive Bayes). We take as a basis for our extension two (Q26 and Q28) of these five queries and extend them to use new types of ML algorithms. Motivated by the recommendations of Singh \cite{singh2016benchmarking}, we create three new workloads (M1, M2 and M3). In this section we describe in detail our workloads.

\subsubsection{Q26} performs a clusterization of customers depending on their book purchasing history. The purchases have to be done in physical stores. 

\subsubsection{Q28} builds a sentiment classifier based on product reviews, to automatically predict if a customer liked a product they purchased based on their review. The underlying classification problem has three different classes - one for positive, neutral and negative sentiment. Based on their written text, each review is assigned to one of the classes. 

\subsubsection{M1}
represents a pattern mining workload, which can identify items that are frequently sold together or categories of products that seem to have a connection. This new workload fits into the business scheme of marketing and enables better product recommendation to the customers. A similar approach is used in BigBench Q1, Q29 and Q30 utilize data mining queries, but they are limited to only pairs of items.

\subsubsection{M2}
is based on Q28 of BigBench V2. We implemented a \acl{LDA} to create a topic model, representing positive, neutral or negative sentiment. Other than Q28 where the algorithm is used to create a classifier to predict the correct sentiment for a given review, this workload analyses the underlying words which make up the review and performs clustering to place each word in the three-dimensional space between the three different sentiments. The idea is based on the assumption, that a review is not purely positive, negative or neutral, but contains parts of each sentiment. 
\subsubsection{M3}
is used for the creation of recommendations of products. It has a similar query to the clustering in Q26, but instead of creating book club buddies based on in-store purchases, the workload tries to predict if a given user is interested in a given item category based on their online browsing history. A shopper is classified as interested in a given category if they have more than average clicks in this item category.
\subsection{Workload Implementation}

This section describes the different workload implementations with their source code available in github \cite{githubCode}. Table \ref{tab:AlgoLibAllWork} illustrates the mapping between workloads, algorithms, and libraries.

\begin{table}
 \caption{Used algorithms and libraries of the workloads. In \textcolor{red}{red} those which are already used in the original BigBench.}
    \centering
    \begin{tabular}{|c|l|c|c|c|c|c|}
\hline
Workload & Algorithm / Library & Mahout & Spark-fim & MLlib & Scikit-learn & SystemML \\ \hline
\multirow{2}{*}{Q26} & K-Means & \textcolor{red}{Yes} & - & Yes & Yes & Yes \\
& \acl{GMM} & - & - & Yes & Yes & Yes \\ \hline
\multirow{3}{*}{Q28} & Naive Bayes & \textcolor{red}{Yes} & - & Yes & - & - \\
& Logistic Regression & - & - & Yes & - & - \\
& \acl{SVM} & - & - & Yes & - & - \\ \hline
\multirow{2}{*}{M1} & Eclat & - & Yes & - & - & - \\
& \acl{FP}-Growth & - & - & Yes & -  & - \\ \hline
M2 & \acl{LDA} & - & - & Yes & -  & -  \\ \hline
\multirow{5}{*}{M3} & Decision Tree & - & - & Yes & - & - \\
& Multilayer Perceptron & - & - & Yes & Yes & - \\
& \acl{SVM}             & - & - & Yes & Yes & Yes \\
& Naive Bayes           & - & - & Yes & - & Yes \\
& Logistic Regression   & - & - & Yes & - & Yes \\ \hline
\end{tabular}
    \label{tab:AlgoLibAllWork}
\end{table}

\subsubsection{Q26} was implemented by using the K-Means algorithm within Mahout. As a new approach, a clustering algorithm based on \acf{GMM} is implemented. The \ac{GMM} algorithm is more complex and assumed to have a higher run time, but since a general spherical distribution is not to be expected from the data, we propose \ac{GMM} to improve the accuracy. Additionally, the algorithm provides information about the probabilities of how likely a chosen point belongs to a particular cluster. To compare the performance of the different libraries both algorithms will be executed by using the following libraries: Mahout, MLlib, Scikit-learn and SystemML.

\subsubsection{Q28} The data used for this workload is first transformed into a \ac{tf-idf}-matrix. This matrix assigns a weight to each word based on its importance. The weight is based on the number of occurrences of the word in a review, with more occurrences leading to a higher value.
The weight accounts for words which often occur across many reviews and is decreased accordingly. Otherwise, words like \textit{and} or \textit{the} - which do not provide significant information gain - would be weighted very high. This is implemented by the inverse document frequency. While the MLlib library provides the tools necessary for this transformation,  SystemML does not. For this reason, the original implementation in Mahout is only compared to the MLlib library, since the transformation is a core part of this workload.

In addition to compare the original two Naive Bayes implementations, Logistic Regression and a \acl{SVM} are also tested on a slightly modified query (the reviews are only classified as positive or negative).

\subsubsection{M1} The \acl{FP}-Growth and Eclat (in its distributed form) algorithms were chosen for implementation. In addition to basic pattern mining, algorithms like \ac{FP}-Growth generate association rules from the observed sets and provide metrics indicating how likely a detected rule is to be true. Eclat used the Spark-fim \cite{Sparkfim} library whereas \ac{FP}-Growth was applied with MLlib.

\subsubsection{M2} The \acf{LDA} is used to create a topic model representing the different sentiments. The workload uses MLlib for executing \ac{LDA}.

\subsubsection{M3} The algorithms chosen for this workload are decision tree classifier and \acl{MLP}. Also, standard classification algorithms (Naive Bayes, Logistic Regression and SVM) are compared to them. This is done to evaluate several implementations for this algorithms across libraries, who are unable to create a \ac{tf-idf}-matrix for the classification task in Q28, as well as comparing the standard classification algorithms with the ones usually used for recommendation. The used libraries are Scikit-learn, SystemML, and MLlib.

\section{Experimental Evaluation} \label{results}
\subsection{Setup and Methodology}
\subsubsection{Hardware}
The experiments were performed on a cluster consisting of 4 nodes connected directly through a 1GBit Netgear switch. All 4 nodes are Dell PowerEdge T420 servers. The master node is equipped with 2x Intel Xeon E5-2420 (1.9GHz) CPUs each with 6 cores, 32GB of main memory and 1TB hard drive. The 3 worker nodes are equipped with 1x Intel Xeon E5-2420 (2.20GHz) CPU with 6 cores, 32GB of RAM and 4x 1TB (SATA, 7.2K RPM, 64MB Cache) hard drives. \cite{ivanov2015performance}

\subsubsection{Software}
All nodes in the cluster are equipped with Ubuntu Server 14.04.1 LTS as the operating system. On top of that the Cloudera Distribution of Hadoop (CDH) version 5.11.0 was installed, which provides Hadoop, HDFS, and Yarn all at version 2.6.0 and Hive 1.1.0. Spark 2.3.0 was installed and Spark SQL was configured to work with YARN and Hive catalog. Library versions used in the setup are Mahout v0.9, MLlib v2.3, SystemML v1.1.0, Python v2.7.6, Scikit-learn v0.20.0 and Pandas v0.19.2.


\subsubsection{Methodology}
All experiments on the different \ac{SF}s were performed three times. The average execution time in seconds of each query and \ac{SF} is published. If during the tests a significant rise of the standard deviation occurred, it is mentioned in the specific execution.

As Scikit-learn can only run on a single node machine, these executions are performed only on the master node of the cluster. All the other libraries are using the whole cluster. Mahout utilizing MapReduce (part of Hadoop YARN); MLlib, Spark-fim, and SystemML running on Spark (on YARN).

\subsection{Data Set Generation and Usage}
BigBench V2 includes a data generator to populate tables and the weblog file with synthetic data, which can be generated in configurable \acf{SF}. Table \ref{tab:datasize} lists the data sizes for the different \ac{SF}s and the workloads for which they are used. The table shows the amount of data processed by the algorithm, not the raw data processed by the query. The weblogs size is also included since it serves as a basis for workloads M1 and M3.

\begin{table}[]
    \centering
     \caption{Overview of the data sizes for different \acl{SF}s}
    \begin{tabular}{|c|c|c|c|}\hline
    Workload & SF\_1 & SF\_10 & SF\_200 \\ \hline
    Q1 / M1 & 1.11 MB & 2.23 MB & 28.17 MB \\ \hline
    Q26  & 189 KB & 336.76 KB & 3.39 MB \\ \hline
    Q28 / M2 & 6.8 MB & 12.48 MB & 132.01 MB \\ \hline
    M3  & 566.16 KB & 997.38 KB & 6.68 MB \\ \hline
    Weblogs & 22 GB  & 40 GB  & 293 GB \\ \hline
    \end{tabular}
    \label{tab:datasize}
\end{table}

\subsection{BigBench V2 ML Workload Results}
\subsubsection{Q26}
\begin{figure}
    \centering
    \includegraphics[width=0.9\linewidth]{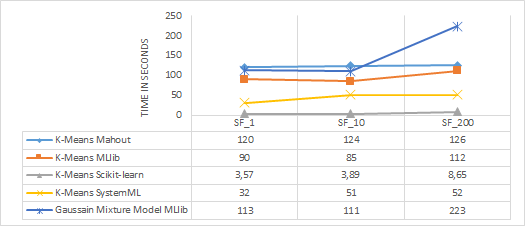}
    \caption{Execution time in seconds on workload Q26 for the different scale factors (SF)}
    \label{fig:exeTimeQ26}
\end{figure}

Figure \ref{fig:exeTimeQ26} depicts the execution time of workload Q26 using the different algorithms and libraries.

It is important to point out, that during multiple test runs of the MLlib K-Means algorithm, a run time variance of about 10\% was observed. This behaviour was not observed at such magnitude by the others libraries.

The Mahout implementation, while being the slowest of all K-Means algorithms in the experiments has a constant run time regardless of the data sizes. The small variation in times are dependant on individual runs and do not differ enough to be counted as actual increase or decrease in run time dependant on the \ac{SF}.

The python implementation with Scikit-learn was the best performing among the algorithms for K-Means, having a significantly lower run time than the others. But compared to the steady time of the mahout algorithm, an increase in run time based on the \ac{SF} could be observed.

While the decrease in run time for the MLlib implementation from \ac{SF} 1 to \ac{SF} 10, can be explained by the general variance in run time, even across multiple tests the run time for \ac{SF} 1 seems to be slower than the times for \ac{SF} 10. Especially when comparing the behavior with other Mllib algorithms there seems to be an underlying pattern, indicating a faster general execution times for \ac{SF} 10 then for \ac{SF} 1.

The tests show that a larger amount of data will be necessary to test these algorithms. We expect that on a large data sizes, the parallel algorithms executed on the cluster will outperform the local python implementation. We plan to address this in the future extensions of the queries.

\subsubsection{Q28}
\begin{figure}
    \centering
    \includegraphics[width=0.9\linewidth]{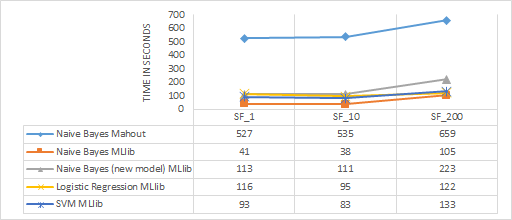}
    \caption{Execution time in seconds on workload Q28 for the different scale factors (SF)}
    \label{fig:exeTimeQ28}
\end{figure}

Figure \ref{fig:exeTimeQ28} reports the results of Q28.
Both algorithms on the original workload show similar behavior, with no significant difference in run time between \ac{SF} 1 and 10 followed by a substantial increase in computational time for \ac{SF} 200. The MLlib implementation is more than 10 times faster for \ac{SF} 1 and 10. However, the difference decreases for \ac{SF} 200, where the Spark run time increases by 181\%, but the Mahout implementation only increasing by 23\%. This could indicate that the Mahout implementation scales better than MLlib, possibly overtaking the Spark implementation for very large data sizes.

Looking at the modified workload, it is remarkable that the algorithms do not seem to be significantly affected by the different number of classes when comparing the values for binary and multiclass classification. The Logistic Regression classifier for the binary problem seems to run a few seconds faster on average than its multiclass implementation, while the Naive Bayes approach observed no significant inherent changes. Logistic Regression starts out as the slowest algorithm for \ac{SF} 1 and 10, but due to better scaling, it is faster than the \ac{SVM} algorithm for \ac{SF} 200.

Both \ac{SVM} and Logistic Regression decrease their run-time performance between SF 1 to SF 10, even though they process a bigger data set. Unlike the K-Means algorithm the run times were much more stable with a difference of under 3\% between test runs, ruling out variance as a factor behind the decrease and further suggesting an increase in parallelism being responsible for the lower time.\\
The fastest algorithm is Naive Bayes, which starts with being twice as fast as the second fastest algorithm. However, Naive Bayes observes the biggest increase from \ac{SF} 10 to 200 with an increase of 171\%, bringing it close to the other two algorithms.

\subsection{New Workloads}
\subsubsection{M1}
First the new algorithms were compared to BigBench V2 Q1 as a baseline. It is a pattern mining workload but limited to only 2 items. Table \ref{tab:exeTimeM3} describes the run times of M1, M2, and M3. 
Even with the higher complexity and the creation of frequent sets of size at least 3, both Eclat and \ac{FP}-Growth achieve faster run times for the first two \ac{SF}s. For the largest \ac{SF} 200, however without drastically increasing the support value, the run time of Eclat increases by more than 100 times, while \ac{FP}-Growth runs out of memory. The much larger item set of \ac{SF} 200 has 100 times more unique items and a much higher amount of transactions. This produces frequent sets containing many items, which is the crucial factor for the much longer run times.
\begin{table}[]
    \centering
    \caption{Execution time in seconds of workloads M1, M2, and M3 for the different scale factors (SF) 1, 10 and 200}
    \begin{tabular}{|c|c|c|c|c|c|}
        \hline
        Workload & Algorithm & Library & SF\_1 & SF\_10 & SF\_200 \\ \hline
        \multirow{3}{*}{M1} & \multicolumn{2}{|c|}{Pattern Mining (Q1)} & 71 & 70 & 75 \\ \cline{2-6}
        & Eclat & MLlib & 43 & 55 & 5791 \\ \cline{2-6}
        & \ac{FP}-Growth & MLlib & 24 & 47 & - \\ \hline
        M2 & LDA & MLlib & 65 & 62 & - \\ \hline
        \multirow{10}{*}{M3} & Decision Tree & MLlib & 75 & 83 & 101 \\ \cline{2-6} 
        & \multirow{2}{*}{Logistic Regression} & MLlib & 78 & 79 & 90 \\ \cline{3-6} 
        & & SystemML & 53 & 53 & 57 \\ \cline{2-6} 
        & \multirow{2}{*}{Naive Bayes} & MLlib & 56 & 58 & 68 \\ \cline{3-6} 
        & & SystemML & 56 & 51 & 52 \\ \cline{2-6} 
        & \multirow{2}{*}{\acl{MLP}} & MLlib & 110 & 121 & 166 \\ \cline{3-6} 
        & & Scikit-learn & 6 & 6 & 9 \\ \cline{2-6} 
        & \multirow{3}{*}{\acl{SVM}} & MLlib & 110 & 110 & 194 \\ \cline{3-6} 
        & & SystemML & 57 & 58 & 72 \\ \cline{3-6} 
        & & Scikit-learn & 29 & 83 & 10102 \\ \hline
    \end{tabular}
    \label{tab:exeTimeM3}
\end{table}
\subsubsection{M2}
\acl{LDA} is used to create a topic model, with a topic representing positive, neutral or negative sentiment.
The execution of the model is not much affected between \ac{SF} 1 and 10. However, when applying the model to the data set with \ac{SF} 200, it terminates with a failure due to a lack of sufficient memory. This raises concerns for the scalability of the algorithm and indicates that it is unsuitable for this data set. It might be possible that the large number of terms (different words) contained in the data causes the problem.

\subsubsection{M3}
Table \ref{tab:exeTimeM3} reports the run times of M3 for the different algorithm regarding the used library and \ac{SF} of the data set.\\
When comparing the \ac{SVM} implementations across libraries, we have the first instance of the locally running Scikit-learn algorithm being overtaken in run time by competitors on the cluster.
While it starts with the lowest run time for the smallest data set, it is slower than the SystemML implementation for \ac{SF} 10 and the most deliberate by a large margin for \ac{SF} 200 with a run time of 2h 48m 22s. The SystemML implementation was overall the fastest and scaled the best among the algorithms.

When comparing Logistic Regression and Naive Bayes classifier, the SystemML implementations are faster than their MLlib counterparts. The behavior of the Logistic Regression algorithms seem similar to a wide gap between the two. Both see a small incline in run time from \ac{SF} 1 to 10 and a more significant incline when scaled up to \ac{SF} 200, with SystemML having the better overall scaling. As for the Naive Bayes implementation, both algorithms start at almost the same time but split from there. SystemML actually decreases in run time, which is the same behavior as the MLlib algorithms for Q28. Only this time there is a definite increase in run time for the MLlib implementation, which increases further for \ac{SF} 200, where SystemML also sees a slight incline, but remains under its original run time for \ac{SF} 1.

When comparing the primarily selected algorithms for this workload, the local implementation of the Scikit-learn \ac{MLP} has the fastest run time, barely seeing an increase for bigger data sets, while the MLlib decision tree is better performing than its \ac{MLP} counterpart. Overall all three algorithms scale well for the task.

\section{Conclusions} \label{confw}

In this paper, we present the results of applying several \acl{ML} algorithms implemented by different libraries on BigBench V2 workloads. 
The experiments have shown that the simple workloads of frequent pattern mining represented in workloads Q1, Q29 and Q30 can be improved by moving on from merely querying the database with more sophisticated machine learning algorithms. 
Workload M1 is a good addition to BigBench V2, representing the common business task of market basket analysis. It can be used as a basis for recommending items to shoppers, or aid in other marketing tasks. In the same notion, M3 was added to predict if a shopper is interested in a given item category based on their online browsing behavior.

During the tests, the original implementations in Mahout had the slowest run times across all workloads, but they seemed to scale well with the growing data sizes. However, it will require a test run with much larger data sizes (in gigabytes) in order to clarify at what scale the distributed cluster algorithms outperforms the local Scikit-learn execution.

SystemML performs best with all cluster based implementations. Even with the problematic format conversion, it still outperforms MLlib, proving that there is a benefit of using the matrix market format as compared to the Spark dataframe model.

In general, the quality of algorithms is not necessarily consistent among one specific library, and a mixture of different implementations might be used to achieve the best results across workloads. In future, we plan to investigate the usage of MLflow \cite{MLflow} or Kubeflow \cite{Kubeflow} to facilitate the integration of the different libraries into the benchmark.\\

\section*{Acknowledgement}
This work has been partially funded by the European Commission H2020 project DataBench - Evidence Based Big Data Benchmarking to Improve Business Performance, under project No. 780966. This work expresses the opinions of the authors and not necessarily those of the European Commission. The European Commission is not liable for any use that may be made of the information contained in this work.


%
%

\bibliographystyle{splncs04}
\bibliography{bibliography.bib}

\begin{acronym}
 \acro{ML}{Machine Learning}
 \acro{GMM}{Gaussian Mixture Model}
 \acro{GEMM}{General Matrix Multiply}
 \acro{FP}{Frequent Pattern}
 \acro{SVM}{Support Vector Machine}
 \acro{RNN}{Recurrent Neural Network}
 \acro{MLP}{Multilayer Perceptron}
 \acro{LDA}{Latent Dirichlet Allocation}
 \acro{DWH}{Data Warehouse}
 \acro{SF}{scale factor}
 \acro{tf-idf}{term frequency - inverse document frequency}
\end{acronym}

\end{document}